# Multiscale Vector-Quantized Variational Autoencoder for Endoscopic Image Synthesis

Dimitrios E. Diamantis
*Department of Computer Science and Biomedical*
*University of Thessaly*
Lamia, Greece
didiamantis@uth.gr

Dimitris K. Iakovidis
*Department of Computer Science and Biomedical*
*University of Thessaly*
Lamia, Greece
diakovidis@uth.gr

***Abstract*—Gastrointestinal (GI) imaging via Wireless Capsule Endoscopy (WCE) generates a large number of images requiring manual screening. Deep learning-based Clinical Decision Support (CDS) systems can assist screening, yet their performance relies on the existence of large, diverse, training medical datasets. However, the scarcity of such data, due to privacy constraints and annotation costs, hinders CDS development. Generative machine learning offers a viable solution to combat this limitation. While current Synthetic Data Generation (SDG) methods, such as Generative Adversarial Networks (GANs) and Variational Autoencoders (VAEs) have been explored, they often face challenges with training stability and capturing sufficient visual diversity, especially when synthesizing abnormal findings. This work introduces a novel VAE-based methodology for medical image synthesis and presents its application for the generation of WCE images. The novel contributions of this work include a) multiscale extension of the Vector Quantized VAE model, named as Multiscale Vector Quantized Variational Autoencoder (MSVQ-VAE); b) unlike other VAE-based SDG models for WCE image generation, MSVQ-VAE is used to seamlessly introduce abnormalities into normal WCE images; c) it enables conditional generation of synthetic images, enabling the introduction of different types of abnormalities into the normal WCE images; d) it performs experiments with a variety of abnormality types, including polyps, vascular and inflammatory conditions. The utility of the generated images for CDS is assessed via image classification. Comparative experiments demonstrate that training a CDS classifier using the abnormal images generated by the proposed methodology yield comparable results with a classifier trained with only real data. The generality of the proposed methodology promises its applicability to various domains related to medical multimedia.***

## I. INTRODUCTION

Disorders of the gastrointestinal (GI) tract constitute significant public health concern, accounting for a considerable proportion of global morbidity and mortality. These conditions not only lead to a reduced quality of life for affected individuals but also impose a heavy financial burden on healthcare systems due to frequent hospitalizations, long-term treatments, and diagnostic procedures [1].

Wireless Capsule Endoscopy (WCE) is a non-invasive screening technique, alternative to Flexible Endoscopy (FE), where a pill-sized camera captures a video of the entire GI tract. Each recording contains >60,000 frames, making manual screening time-consuming (45–90 minutes) and prone to human error. To address this, Clinical Decision Support (CDS) tools based on image analysis have been developed [2–4]. Their effectiveness depends on generalization, and as such they require large and diverse datasets [2]. However, medical data are relatively hard to obtain due to privacy regulations such as GDPR [5], as well as the cost and annotation effort [2]. Public WCE datasets remain scarce and are often imbalanced with low diversity [6], limiting the performance of deep learning-based CDS systems.

In the context of endoscopic image synthesis, studies indicate that generating high-quality synthetic images continue to pose significant challenges [7–10] such as plausibility issues and image artifacts, both of which are primarily linked to the availability of diverse training samples [11]. Current GAN-based methodologies struggle to capture the natural features of the GI tissue, *i.e.* natural structure, texture and color [8], [10], especially on abnormality generation tasks due to the visual diversity of the different lesions, such as polyp types and inflammation conditions. Consequently, much work focuses on polyp synthesis [7], [8], [10–13] but seamlessly integrating generated pathologies into the background remains a challenge.

Recently Variational Autoencoders (VAEs) [14] have been successfully used in the context of small bowel image synthesis targeting inflammatory conditions [15], [16] generating more diverse and plausible images compared to GAN-based methodologies. While the quality and diversity of those methodologies outperformed GANs in the context of WCE image synthesis, their diversity performance is affected by their small, continuous latent representation. Furthermore, while the images generated were plausible, their resolution was limited (i.e. 96×96 pixels in size) and limited to generating specific types of abnormalities, i.e. mucosal inflammations.

Aiming to cope with these issues, in this paper we propose a novel Vector Quantized Variational Autoencoder (VQ-VAE) [17], that generates high resolution, realistic synthetic small bowel WCE images with different types of abnormalities that include mucosal inflammations, vascular conditions and polyps, which are more frequently observed in patient populations. The contributions of this work can be summarized as follows:

This work is part of the European project SEARCH, which is supported by the Innovative Health Initiative Joint Undertaking (IHI JU) under grant agreement No. 101172997. The JU receives support from the European Union's Horizon Europe research and innovation programme and COCIR, EFPIA, Europa Bio, MedTech Europe, Vaccines Europe, Medical Values GmbH, Corsano Health BV, Syntheticus AG, Maggioli SpA, Motilent Ltd, Ubitech Ltd, Hemex Benelux, Hellenic Healthcare Group, German Oncology Center, Byte Solutions Unlimited, AdaptIT GmbH. Views and opinions expressed are however those of the author(s) only and do not necessarily reflect those of the aforementioned parties. Neither of the aforementioned parties can be held responsible for them.

- It proposes a methodology that enables the seamless introduction of abnormalities into normal WCE images using the proposed MSVQ-VAE.
- It proposes a novel extension of the VQ-VAE architecture, named MultiScale Vector Quantized Variational Autoencoder (MSVQ-VAE) that makes use of multiscale feature extraction and residual connections, that capture multiscale features at three different scales forming a feature-rich representation codebook. To the best of our knowledge in the context of VQ-VAE this combination has not been previously considered.
- It enables conditional generation of high-quality synthetic images allowing the model to introduce specific types of abnormalities into normal WCE images based on given conditions.
- It demonstrates the effectiveness of the model across a variety of abnormality types, including polyps, vascular and inflammatory conditions.

The performance of the proposed methodology is evaluated in the context of WCE abnormality detection showing that the resulting images can be used to effectively train CDS systems to identify abnormalities in real images.

The rest of this paper consists of five sections. In Section II related work in the context of medical image generation is presented. In Section III the architecture of the proposed MSVQ-VAE is described in detail alongside the proposed methodology. The evaluation methodology and the results obtained are described in Section IV while the conclusions derived from this study along with the perspectives for future work are summarized in the last section.

## II. Related Work

Synthetic medical image generation, particularly using Generative Adversarial Networks (GANs), has attracted significant research attention across various applications. Many prominent GAN architectures have been explored for medical image synthesis. For example, adaptations of Pix2Pix produced chest X-rays with nodules [18] and synthesized 3D CT nodules [19]. Progressive Growing GAN (PGGAN) created dermoscopic images [20], and a multiscale GAN fused MRI modalities for enhanced diagnostic value [21]. CycleGAN [22] was employed for unpaired MRI-CT brain image translation [23] and modified for 3D cardiac/pancreatic synthesis and segmentation [24].

Generating realistic endoscopic images is considered more difficult [11], [25], [26], as these images often lack the defined patterns seen in modalities like CT or MRI [11]. Several GAN-based methods have been applied, often focusing on polyp synthesis. These include GANs conditioned on edge-filtering/masks [10], inpainting methods that add polyp features to normal images [27], patch-based methods for gastric findings [8], potentially requiring manual placement [7], and CycleGAN trained with random polyp masks [13]. Variations enhanced simulator images [28] or used dual GANs [12]. However, synthesis quality in some cases [12], [13] depended on mask positioning. StyleGANv2 [29] yielded realistic polyp images [11] but required a large private dataset, limiting reproducibility. Recently [30] a mask-conditioned based diffusion model was proposed to synthesize high-quality endoscopic images with polyps using polyp masks with promising results. Stable Diffusion [31] based models have also been employed to generate realistic endoscopic images. Notably [32] proposed a method for image-to-image translation based on fine-tuning a stable diffusion model and controlling its output using a ControlNet [33], that generated realistic images from synthetic 3D models. The model was trained on just 100 real images and was able to generate realistic results. However, despite the low data requirement, the approach has notable limitations, including poor temporal consistency and a high sensitivity to the choice of training samples.

Fewer studies focus on WCE image generation. WCE presents unique challenges like lower resolution and fewer abnormal images due to the inability to control the capsule, unlike FE [6]. WCE is vital for examining the small bowel, especially for IBD/Crohn's disease [34], although small bowel neoplasia is less common than in the colon or stomach [35]. Specific WCE generation efforts include adapting a hybrid VAE-GAN for data augmentation [36], though synthetic images were somewhat blurry. A Texture Synthesis GAN (TS-GAN) aimed to synthesize inflammatory conditions [37] but produced artifacts, possibly due to scarce training data. Subsequently, EndoVAE, a conventional VAE with single-scale filters, was proposed [15]. Similarly, an improved version of EndoVAE was proposed with named TIDE [16] which employees multiscale feature extraction in both the encoder and decoder part to generate higher quality, more diverse WCE images..

This study presents a novel VQ-VAE architecture named MultiScale Vector Quantized Variational Autoencoder (MSVQ-VAE) that makes use of multiscale feature extraction and residual connections to generate high resolution, plausible WCE images with abnormal findings from three conditions: vascular irregularities, mucosal inflammation, and polypoid lesions.

## III. Proposed Methodology

### A. Architecture

The proposed MSVQ-VAE architecture can be broken down into three main components: a) the encoder, b) the vector quantization layer and c) the decoder. Key components of the encoder and the decoder networks of architecture are the multiscale feature extraction modules (MSB) which capture a feature-rich representation of the input volume along with residual connections that help propagate the input volume representation across the MSB modules.

The MSB module (Fig. 1) aims to capture a feature-rich representation of the input volume using multiscale feature extraction. More specifically using convolutional layers the module extracts features under three different scale, i.e. small, medium and large with 3×3, 5×5 and 7×7 filter size respectively with a stride of 1. The output of those convolutions is then concatenated across the depth dimension forming a volume of distinct feature maps. The concatenated volume has three times the number of feature maps of the original input volume. This is then passed by two consecutive convolutional layers whose role is to aggregate the feature maps across the depth dimension and perform feature-map reduction acting as a bottleneck for the module. The aggregation of the features is handled by a pointwise convolution [38] while the feature map reduction is handled by a convolutional layer with 3×3 filters and a stride of 1. All the activation functions of the convolutional layers of the module are Rectified Linear Units (ReLU).

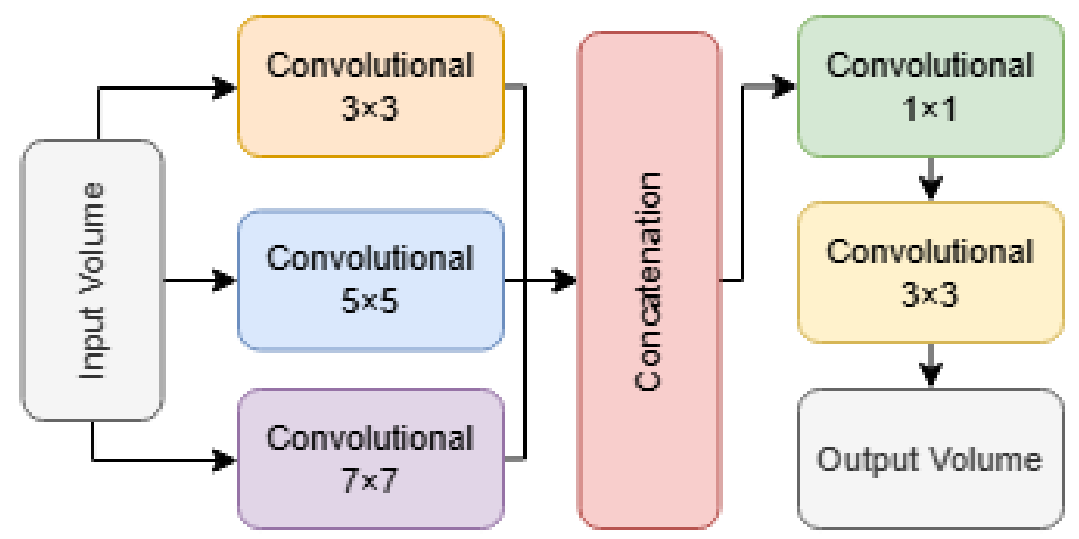

Fig. 1. The architecture of the Multiscale Feature Extraction module (MSB).

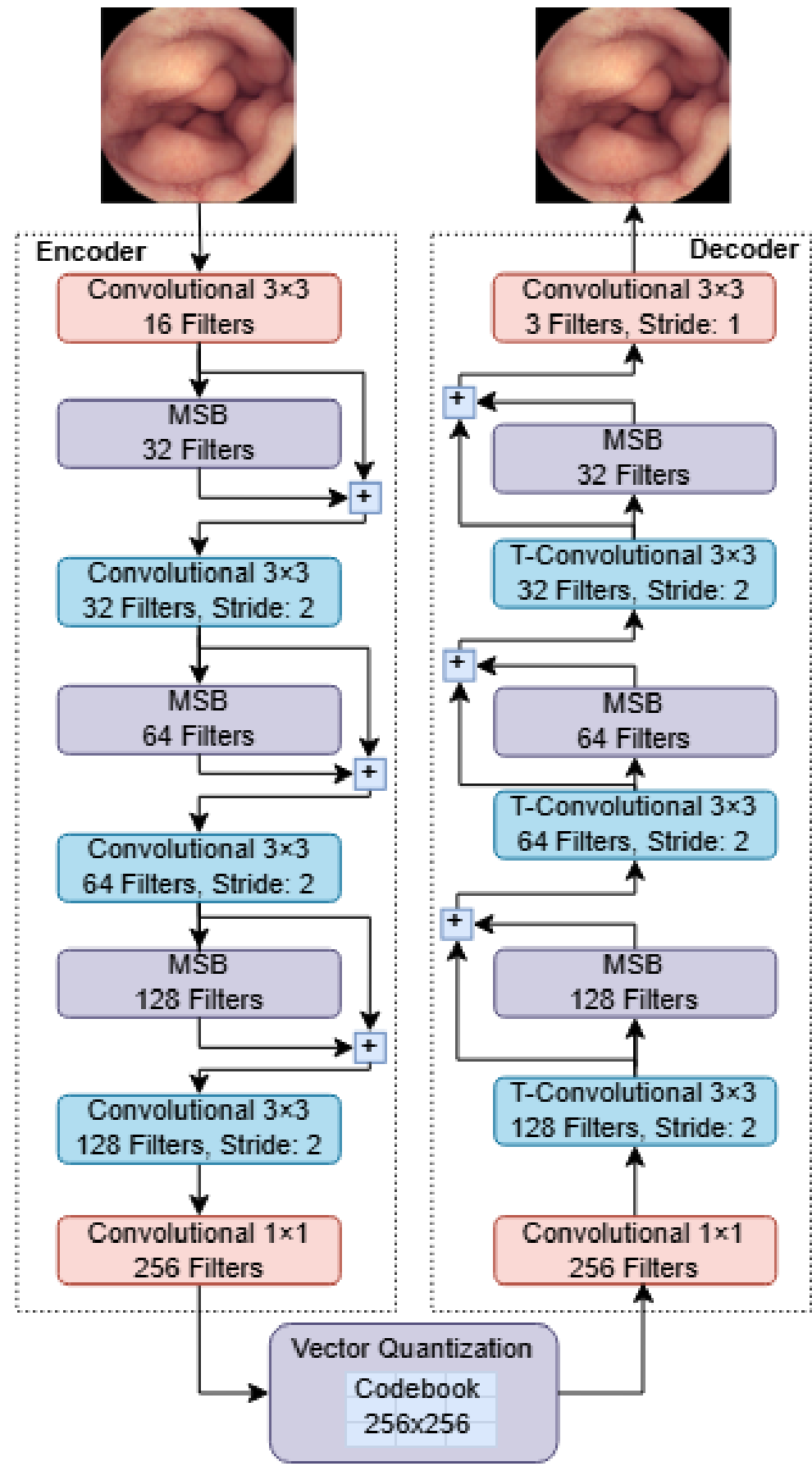

Fig. 2: The architecture of MSVQ-VAE

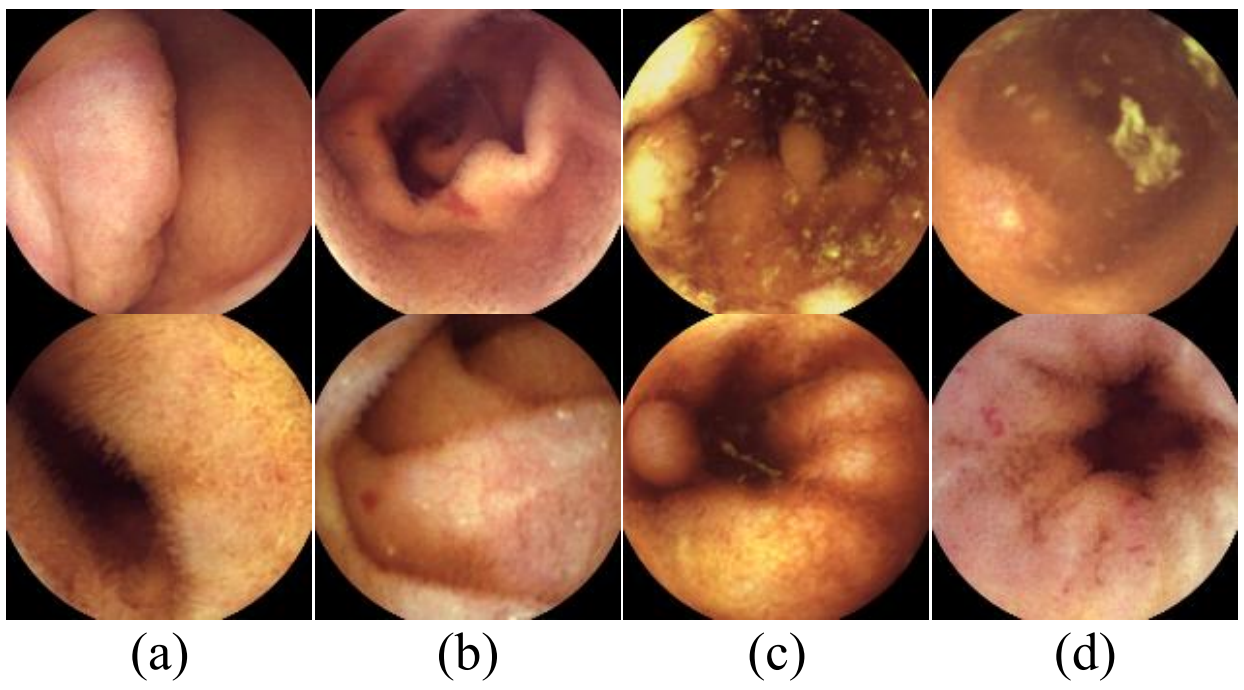

Fig. 3. Sample images from the KID [42] WCE image database. (a) Healthy small bowel tissue, (b) Vascular lesions, (c) Polyps, (d) Inflammations.

The proposed MSVQ-VAE architecture is illustrated in Fig. 2 and it is composed of two networks; the encoder and the decoder. The encoder network receives as an input an RGB image of 224×224 pixels in size and is passed through a convolutional layer with 3×3 filters and a stride of 1 extracting 16 feature maps. These feature maps are then followed by three series of MSB modules and convolutional layers with stride of 2, each one reducing the spatial dimensionality of the input volume in half and doubles number of extracted feature maps. Similarly to the encoder, the decoder network follows the reverse procedure, *i.e.,* it is composed of a series of MSB modules and transposed convolutional layers, each one doubling the spatial dimensionality and decreasing the number of features by half. The final layer of the decoder consists of a convolutional layer with filters 3×3, stride of 1 and 3 feature maps, each one representing a channel of the reconstructed RGB image. While multiscale feature extraction has been employed by many architectures such as Inception-based [39], they typically extract features under two scales and focus primarily on feature map number reduction aiming to reduce the computational complexity of the network. Contrary to that, the MSB module aims to capture a feature-rich representation of the input volume, extracting the same number of features under all scales. This is important, especially in the case of training the architecture with relatively small datasets with little diversity. The feature map number reduction is handled at by the bottleneck layer later in the MSB module.

Aiming to battle the problem of vanishing gradient while training and to preserve higher level features across the MSB modules, both the encoder and decoder networks employee residual connections. More specifically the input of each MSB module pass through a convolutional layer which extracts the same number of features as its input with filter size of 3×3. The output of this layer is then aggregated using the algebraic sum operation to the output feature maps of the MSB module, followed by a pointwise convolution. While nowadays residual connections are common a common architectural component of deep networks, they typically skip a single layer, focusing primarily on battling the vanishing gradient problem. Contrary to that, the residual connections of MSVQ-VAE architecture skip the entire MSB module which is composed of multiple convolutional layers which along with battling the problem of vanishing gradients, preserve higher level features across the network.

The latent representation of the proposed architecture is handled by a vector quantization layer [17], implemented using an Exponential Moving Average (EMA) codebook update strategy. Given a continuous encoder output $z_e(x) \in \mathbb{R}^D$,where $x$ is the input image and $D$ is the dimensionality of the latent space, the layer discretizes the representation by selecting the nearest codebook vector from a learned set of embeddings $\mathcal{E} = \{e_k \in R^D\}_{k=1}^{K}$, where $K$ is the total number of codebook entries and $e_k$ denotes the $k$-th embedding vector. The quantized latent vector is obtained via nearest-neighbor lookup: $z_q(x) = e_{k^*}$ where $k^* = \arg\min_k ||z_e(x) - e_k||_2$ , and $||\cdot||_2$denotes the Euclidean norm.

Unlike the original VQ-VAE formulation, where embeddings are updated using the gradient descent, MSVQ-VAE employs exponential moving averages for codebook updates, which are more stable and non-parametric. This strategy avoids codebook collapse and noisy updates that may arise from infrequent or sparse usage of embeddings in mini-batches, which is more evident in small, non-diverse datasets. In effect, EMA decouples codebook updates from gradient backpropagation ensuring that each embedding evolves based on accumulated statistics, leading to more reliable convergence and better utilization of the latent space [17].

Let $N_k$ denote the number of encoder outputs assigned to code $e_k$ in a training batch, and let $m_k \in \mathbb{R}^D$ be the sum of encoder outputs assigned to $e_k$. The running estimates $\widehat{N_k}$ and $\widehat{m_k}$ are updated as follows:

$$\widehat{N_k} \leftarrow \gamma\widehat{N_k} + (1-\gamma)N_k \tag{1}$$

$$\widehat{m_k} \leftarrow \gamma\widehat{m_k} + (1-\gamma)m_k \tag{2}$$

$$e_k \leftarrow \frac{\widehat{m_k}}{\widehat{N_k} + \epsilon} \tag{3}$$

where $\gamma \in [0,1)$ is the EMA decay factor controlling the update speed, and $\epsilon > 0$ is a small constant used to avoid division by zero. Since the embeddings are updated separately from the gradient flow, the loss function omits the embedding loss and consists of two terms: a reconstruction loss and a commitment loss. The reconstruction loss is computed as the mean squared error (MSE) between the input and the reconstructed output:

$$\mathcal{L}_{rec} = \frac{1}{N}\sum_{i=1}^{N} | x_i - \widehat{x_i}|^2 \tag{4}$$

where $N$ is the number of pixels in the image, $x_i$ is the value of the $i$-th pixel in the original input, and $\widehat{x_i}$ is its reconstructed counterpart. The total loss is then given by:

$$\mathcal{L} = \mathcal{L}_{rec} + \beta||z_e(x) - \mathrm{sg}[z_q(x)]||^2 \tag{5}$$

where $\beta$ is a weighting factor for the commitment loss, and sg[·] denotes the stop-gradient operator that blocks gradient flow through its argument. This formulation provides a stable and expressive discretization mechanism, enabling MSVQ-VAE to learn compact latent representations that are well-suited for high-fidelity generation and compression tasks.

### B. Methodology

Generating novel images with a VQ-VAE necessitates explicitly modeling the prior distribution, over its learned discrete latent codes. While this architecture effectively compresses images into sequences of codebook indices via the encoder and reconstructs them using the decoder, it does not inherently learn the statistical dependencies determining which code sequences correspond to plausible images. To enable generation, a secondary model must be trained to capture this prior knowledge. Autoregressive models, such as PixelCNN [40], or related sequence regressors, are commonly employed for this task, learning to predict likely sequences of latent codes. Image synthesis then proceeds by sampling codes from this trained prior and subsequently using the VQ-VAE decoder to render the final output.

While training a regressor on VQ-VAE codes effectively generates plausible images from large, diverse datasets like natural images, this approach struggles with limited datasets. More specifically, in the context of WCE image synthesis, the scarcity of training samples often results in the regressor producing unrealistic images that fail to capture the small bowel natural structure or more often preserve the specific characteristics of desired anomalies. Aiming to combat these drawbacks, the proposed methodology makes use of real images to guide the MSVQ-VAE decoder to generate high quality WCE images with abnormalities instead of the conventional autoregressive approach.

Novel abnormal images are generated using the trained MSVQ-VAE architecture through a four-step process: a) initially a normal and an abnormal image are encoded using the encoder network of the architecture passing and their corresponding discrete latent codes are obtained from the vector quantization layer of the architecture; b) then the binary mask of the abnormal image is used to obtain the spatial location of the abnormality within the discrete latent codes of the abnormal image; c) the discrete latent codes of the abnormal image are then used to replace the latent codes of the normal image using the obtained spatial location of the abnormality, forming a new index map representation; d) finally, the index map is then decoded and forwarded to the decoder which handles the generation of the novel abnormal image. The characteristics of the abnormality are integrated and seamlessly blended within the surrounding features of the normal image directly by the decoder, primarily due to the multiscale feature extraction process followed by the MSB modules.

## IV. Experiments and Results

### A. Dataset and Experimentation Setup

To evaluate the performance of the proposed MSVQ-VAE architecture in the context of WCE image synthesis, we used the KID [41] image database (Fig. 3). The dataset consists of 2.352 RGB images of 360×360 pixels in size from which 1.778 images represent healthy GI tissue while the rest 574 images contain a variety of pathological findings found along the GI tract, including vascular (303 images), polyps (44 images) and inflammatory (227 images) conditions, along with their corresponding binary masks. To train the MSVQ-VAE, we selected 728 normal images from the small bowel, and all the available abnormal findings. To use the selected WCE images for training, the images were cropped to size 320×320 pixels to remove the surrounding borders and resized to 224×224 pixels. A series of experiments were performed to select the optimal hyper parameters of the proposed MSVQ-VAE architecture. The experiments focused primarily on the selection of the embedding dimensions $D$, the number of embeddings $K$ and codebook commitment $\beta$. In each test we evaluated the codebook usage starting with $\beta = 0.1$ and $K = 32$ and $D = 16$. We found that the optimal commitment was $\beta = 0.25$ and $K, D = 256$ as that enabled the architecture to make use of the entire codebook. Large embedding dimensions or number of embeddings significantly affected the codebook usage, *i.e.* when $K = 512$ and $D = 256$ the codebook usage dropped to 82% while similarly when $K = 256$ and $D = 512$ the codebook utilization did not surpass 61%. To train the proposed architecture we used the Adam optimizer with initial learning rate set to 0.001 with an exponential decay of 0.9. The network was trained for 2.000 epochs and used a batch size of 64 with random horizontal and vertical flips as data augmentation on the training images.

### B. Results

Following the steps described in Section 3.2 the trained MSVQ-VAE architecture was used to generate abnormal images by randomly selecting and combining pairs of normal and abnormal images from the KID dataset. Smaller abnormalities with a spatial dimension of ≤20% of the total image size were selected because they are usually more challenging and crucial to detect, *e.g.*, small polyps may evolve into malignancies if not detected early. Therefore, all experiments were conducted within this constraint. Figure 4 illustrates indicative samples of normal images along with respective generated images with seamlessly blended abnormalities.

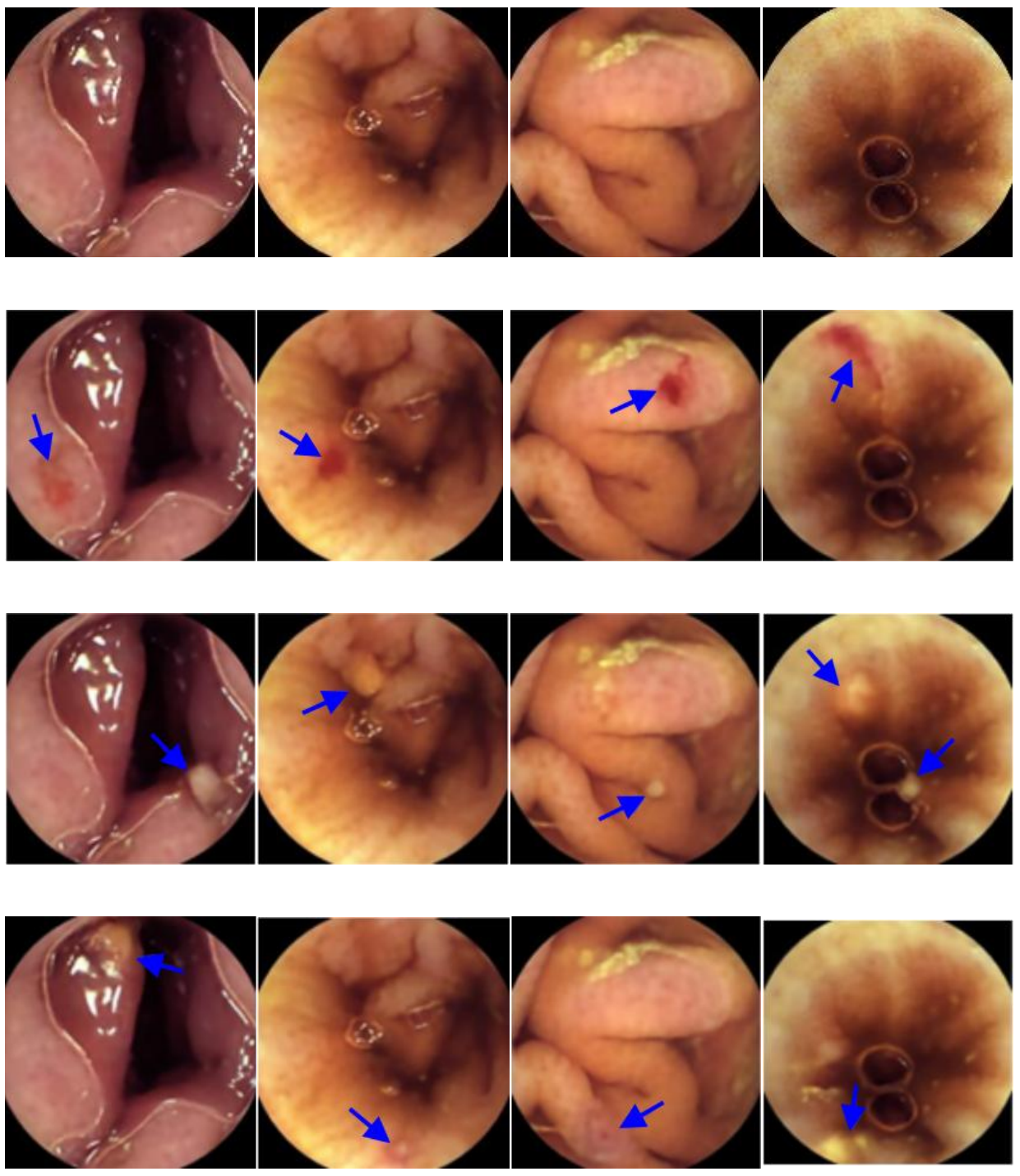

Fig. 4: Sample images generated using the proposed MSVQ-VAE architecture. The first row illustrates real normal images from the KID [42] WCE image database and the next generated images containing vascular lesions (second row), polyps (third row) and inflammations (last row). The blue arrows point to the location of each abnormality.

To evaluate the plausibility and utility of the synthetic abnormal images, a state-of-the-art classifier for WCE CDS [42] was trained using the real normal small bowel WCE images from the KID dataset and the abnormal images generated by the proposed architecture. The classifier was tested on the real abnormal images of the dataset. Then the same classifier was trained using real normal and real abnormal images, and the results of the two experimental setups were compared. To ensure a fair comparison, we followed a 5-fold cross validation procedure where in each fold the classifier was tested only on real abnormal images that were not used as base anomalies for the training set. Furthermore, as there is severe class imbalance in the dataset, for the comparative experiment we limited the number of training abnormal samples to match the proportions of the real dataset. Given this class imbalance, the Area Under the ROC Curve (AUC) was selected as the performance measure. AUC is recognized for being reliable, intuitive, and, importantly, insensitive to such imbalances [43]. The results obtained from the experiments show that although the classifier achieves higher classification (AUC 84.3±1.4%), when trained on real images it achieves a slightly lower performance (AUC 79.8±1.8%) when the classifier is trained with synthetic abnormal images and tested on real ones. This indicates that the resulting images are plausible and can be used to enhance the classification performance or balance imbalanced datasets such as the one used in this study.

## V. Conclusions and Future Work

This paper presented MSVQ-VAE a novel extension of the VQ-VAE architecture that makes use of multiscale feature extraction and residual connections. Unlike other VAE-based SDG models for WCE image generation it is used to seamlessly introduce abnormalities into normal WCE images. In addition, it introduces a methodology for conditionally generating synthetic images, allowing specific types of abnormalities to be incorporated into normal WCE images. Through experimentation we presented that architecture can capture the complex structure of the small bowel lining and successfully combine it with abnormalities of three different conditions: vascular irregularities, mucosal inflammation and polypoid lesions.

Compared to methods such as [10] and [27], MSVQ-VAE does not rely on edge filtering or complex pipelines to incorporate abnormalities in normal WCE images. Such multi-stage processes can introduce visual artifacts and unnatural boundaries, compromising the plausibility of the synthesized images. As such, the direct approach followed by MSVQ-VAE minimizes potential artifacts introduced by pre-processing steps which ensures that abnormalities are seamlessly blended by the decoder. Furthermore, while other methods are limited to polyps, MSVQ-VAE can introduce more complex anomalies, including vascular lesions and inflammations which exhibit significantly larger textural and color variations compared to polyps.

Comparative experiments were conducted, between two state-of-the-art classifiers; one trained and tested on real images and one trained on real normal and synthetic abnormal images and evaluated on real data. The results demonstrate that the proposed architecture can effectively generate highly plausible images for balancing imbalanced datasets, such as the one used in this study. Furthermore, the experiments showed that MSVQ-VAE can generalize well even with relatively small and class-imbalanced training sets.

Future work includes the investigation of attention-based methods which could enhance the performance of the MSB module on the representation and blending of the abnormalities into normal images, which may also enable the increase of embedding dimensions positively, affecting the quality of the overall image synthesis. A limitation of the current methodology is the requirement for a mask to guide the placement of the abnormality within the GI lining. Future efforts will focus on automating this process to improve the practicality of the model. Considering the generality of the proposed methodology, its application on different domains related to medical multimedia is also a promising perspective to be investigated.